\documentclass{article} % For LaTeX2e
\usepackage[final]{acl}

\usepackage{amsmath,amsfonts,bm}

\def\eqref#1{equation~\ref{#1}}
\def\1{\bm{1}}

\DeclareMathAlphabet{\mathsfit}{\encodingdefault}{\sfdefault}{m}{sl}
\SetMathAlphabet{\mathsfit}{bold}{\encodingdefault}{\sfdefault}{bx}{n}

\usepackage{times}
\usepackage{hyperref}
\usepackage{url}

\usepackage{algorithm}
\usepackage{algorithmic}
\usepackage{inconsolata}

\title{\textsc{MAQuA}: Multi-outcome Adaptive Question-Asking for Mental Health using Item Response Theory}

\author{
Vasudha Varadarajan\textsuperscript{$\spadesuit$}\thanks{equal contribution},
Hui Xu\textsuperscript{$\heartsuit$}$^*$,
Rebecca Astrid Boehme\textsuperscript{$\dagger$},\\
\textbf{Mariam Marlan Mirström\textsuperscript{$\ddagger$},
Sverker Sikström\textsuperscript{$\ddagger$} \&
H. Andrew Schwartz\textsuperscript{$\heartsuit,\diamondsuit$}}\\
 \textsuperscript{$\spadesuit$}Language Technologies Institute, Carnegie Mellon University.
\\
\textsuperscript{$\heartsuit$}Stony Brook University, 
\textsuperscript{$\dagger$}Aarhus University, 
\textsuperscript{$\ddagger$}Department of Psychology, Lund University.\\
\textsuperscript{$\diamondsuit$}College of Connected Computing, Vanderbilt University. 
}

\usepackage{microtype}
\usepackage{graphicx}
\usepackage{subfigure}
\usepackage{booktabs} % for professional tables

\usepackage{hyperref}

\usepackage{amsmath}
\usepackage{amssymb}
\usepackage{mathtools}
\usepackage{amsthm}
\usepackage{makecell}
\usepackage[capitalize,noabbrev]{cleveref}

\theoremstyle{plain}

\theoremstyle{definition}

\theoremstyle{remark}

\usepackage[textsize=tiny]{todonotes}

\begin{document}
\maketitle
\begin{abstract}
Recent advances in LLMs offer new opportunities for scalable, interactive mental health assessment, but excessive querying burdens users and is inefficient for real-world assessments across transdiagnostic symptom profiles.
We introduce \textsc{MAQuA}, a multi-outcome modeling and adaptive question-asking framework for simultaneous, multidimensional mental health assessment. Combining multi-outcome modeling on language responses with item response theory (IRT) and factor analysis, \textsc{MAQuA} selects the questions with most informative responses across \textit{multiple} dimensions at each turn to optimize diagnostic information, improving accuracy and potentially reducing response burden.
Empirical results on a novel dataset reveal that \textsc{MAQuA} reduces the number of assessment questions required for score stabilization by 50–87\% compared to random ordering (e.g., achieving stable depression scores with 71\% fewer questions and eating disorder scores with 85\% fewer questions). \textsc{MAQuA} demonstrates robust performance across both internalizing (depression, anxiety) and externalizing (substance use, eating disorder) domains, with early stopping strategies further reducing patient time and burden.
These findings position \textsc{MAQuA} as a powerful and efficient tool for scalable, nuanced, and interactive mental health assessment, advancing the integration of LLM-based agents into real-world clinical workflows.
\end{abstract}

\section{Introduction}

Recent progress in large language models (LLMs) has enabled the automated inference of mental health scores from patient-generated natural language.
However, comprehensive evaluations indicate that such LLM-based assessments are inconsistent~\cite{ji2022mentalbert} and, in many cases, less accurate than dedicated, condition-specific models with established psychometric validity~\cite{harrigian2020models}. These limitations present critical barriers to the clinical adoption and trustworthiness of LLMs in the mental health domain.

Traditional NLP approaches in mental health have often relied on annotations of specific conditions derived from social media, primarily focusing on single-task models, such as encoder-based classifiers for depression~\cite{coppersmith2015clpsych,eichstaedt2018facebook}, suicidal ideation or risk~\cite{shen2017depression, varadarajan-etal-2024-archetypes} and anxiety~\cite{owen2020towards,gkotsis2017characterisation, juhng2023discourse}, detection. %which encode a single dimensional mental health score. %More recently, advances with LLMs (e.g., Mental LLaMA, Mental LLM) have enabled the community to explore one model
The narrow scope of modeling a single dimensional mental health score typically fails to capture comorbidities or the complex, multidimensional nature of symptoms observed by real-world clinicians~\cite{shani-stade-2025-measuring, soni-etal-2025-mental, varadarajan-etal-2025-linking}.
% methods typically provide static, single-dimensional measurements of mental health states and rarely capture the complex, multidimensional symptom profiles encountered in clinical practice~\cite{transdiagnostic_opinion_paper_betsy}
% Static, single-dimensional measurements of mental health states from social media are common~\cite{}; There is scarce exploration of multidimensional models suitable for capturing the complex interplay of symptoms encountered in clinical practice~\cite{}. 
More importantly, they do not address the interactive paradigm where LLM agents engage with users in \textit{prompted} settings: language generated in response to structured diagnostic interviews, as would be typical in real-world clinical settings.

% While large language model (LLM)-based agents %are rapidly transforming mental health applications through their advanced natural language capabilities and widespread deployment in clinical and diagnostic settings. 
% %These models 
% hold significant promise for enhancing traditional assessment and intervention paradigms by offering scalable, accessible, and interactive mental health support~\cite{hua2025scoping}, challenges such as hallucinations, inconsistent or inappropriate responses, and uncertainties about appropriate behavior in sensitive clinical contexts pose substantial risks to their safe and effective use~\cite{lawrence2024opportunities, ji??}.
% To mitigate these risks, integrating complementary non-LLM methods: such as rule-based systems, clinical decision support as well as psychometrically validated tools can create models that leverage both LLM conversational strengths and clinically grounded frameworks,
% thereby improving diagnostic accuracy while maintaining clinical validity and enhancing trustworthiness and interpretability in mental health applications.

% Additionally, clinicians intuitively adapt their queries based on information obtained earlier in the interaction-avoiding redundancy, probing ambiguity, or targeting emergent concerns: while LLMs, though adept at modeling linguistic associations, may fail to dynamically ground inferred states to underlying mental health constructs.
Further, in actual clinical practice, clinicians dynamically adapt their lines of questioning based on prior information received, avoiding redundancy, clarifying ambiguous responses, and addressing emergent concerns~\cite{james2010science, welch2025isca}.
While LLMs excel at modeling linguistic patterns, there are mixed signals on their ability to dynamically ground inferred states in underlying mental health constructs~\cite{singh-etal-2025-systematic, ganesan2024explaining}, especially given the multi-objective challenge of simultaneously selecting diagnostic questions and evaluating mental health status~\cite{li2025aligning, sener2018multi}. Furthermore, maximizing the diagnostic yield within the constraints 
of limited clinician-patient interaction time remains a key priority, as excessive probing, especially via LLM-based dialogue agents can be mentally taxing and lead to decision fatigue or disengagement~\cite{jin2025don}, highlighting the need for adaptive systems that select only the most informative and relevant follow-up questions.
%This highlights the need for adaptive, contextually aware and nuanced assessment tools for complementing mental health dialogue agents. 

%While current efforts predominantly focus on internalizing conditions, thoughts and emotions that are expressed via language, like depression or suicidal ideation~\cite{}, 
 %there is less attention paid to externalizing disorders: observable behaviors such as aggression, substance misuse, or risk-taking are often elicited through targeted questions rather than passive monitoring. 
 %This highlights the need for comprehensive, dialogue-driven assessment frameworks that can precisely estimate multiple mental health conditions. 
 %Our approach explicitly integrates diverse mental health constructs to enable simultaneous, nuanced modeling across multiple clinical dimensions. 
 To address these challenges, we propose \textsc{MAQuA}: an adaptive, language-based assessment framework that supports multidimensional mental health modeling. This framework infers multiple underlying condition scores while adaptively selecting the most informative follow-up questions, thereby guiding interactions efficiently toward richer diagnostic insight, making it suitable to operate alongside LLM agents. Building on item response theory (IRT)-based adaptive assessments introduced by ~\citet{varadarajan-etal-2024-alba}, our results demonstrate that optimizing information gain  across multiple conditions simultaneously can be even more effective than single-score models.

To explore this, we empirically benchmark single-task and
multitask models
on a multidimensional mental health dataset and assess the effectiveness of adaptive question selection. 
Our research investigates (1) whether cross-condition information sharing improves per-condition predictive performance; 
(2) validity of automatically inferred mental health dimensions; and 
(3) the efficacy of multidimensional IRT in sustaining validity of measures in subsequent adaptive assessment turns. Our main contributions include: 
(a) a systematic comparison of multi-condition modeling techniques, 
(b) an adaptive assessment framework for optimizing information gain across outcomes, 
(c) empirical results demonstrating robust gains in multidimensional assessment, and 
(d) the release of a novel, questionnaire-driven dataset to support future research.

\section{Related Work}

%Our research addresses a critical need for mental health assessments that are simultaneously efficient, accurate, and comprehensive enough to reflect complex symptom profiles of individuals. 
%As part of efforts towards building scalable, efficient and accurate AI for mental health, our work builds on three convergent lines of research: language-based assessment, adaptive testing, and multitask learning.

% Language-Based Assessment

While large language models (LLMs) have shown promising results in zero-shot and few-shot prediction tasks~\citep{ganesan2024explaining, hur2024language}, finetuned or instruction-tuned models remain generally more reliable and better validated across a range of mental health outcomes~\citep{xu2024mental}. Recent advances demonstrate that open-ended language responses to standardized questions can predict mental health scores with high accuracies: sometimes with correlations exceeding 0.8 with established clinical rating scales~\citep{kjell2022natural,varadarajan-etal-2024-alba,sikstrom2023precise}. These include pre-trained language models tailored for mental healthcare applications, such as ClinicalBERT and MentalBERT~\citep{alsentzer2019publicly,ji2022mentalbert}.

Large language models (LLMs) commonly address referential and vague queries by employing targeted, selective prompts, which has been shown to improve answer accuracy and reduce errors~\cite{zhang2025modeling, kuhn2022clam}. However, despite these improvements, LLMs still fall short of human conversational subtlety and adaptiveness when it comes to clarification and follow-up questions. In terms of LLMs for mental health support, \citet{rosenman-etal-2024-llm} demonstrate that LLMs can effectively transform unstructured psychological interviews into structured questionnaires, enabling automated, multidimensional psychiatric evaluation, though reliability and consistency still require further improvement for clinical deployment. Complementing this, \citet{nguyen-etal-2025-large} explore LLMs’ ability to engage in mental health counseling, showing that, though LLMs can generate contextually relevant follow-up questions, they often lag behind human clinicians in empathy, specificity, and diagnostic nuance, and in crafting clarifying or probing questions that are crucial for effective counseling. Similarly, \citet{yang-etal-2023-towards} assess LLM performance across a spectrum of mental health tasks, finding that LLMs frequently overlook emotional cues or oversimplify questions, limiting their utility for nuanced clinical interpretation. This underscores the need for methods like ours that explicitly model multiple mental health factors to guide strategic question selection. Our framework enhances both the efficiency and accuracy of mental health assessments by optimizing inquiry and serving as a comprehensive diagnostic tool.

%Alongside innovations in response analysis, adaptive testing has made assessment dramatically more efficient. Computerized adaptive testing (CAT), grounded in Item Response Theory (IRT)~\citep{embretson2000item,hambleton1991fundamentals}, replaces lengthy fixed interviews with a few tailored questions. This approach can reduce the number of administered items by over 50\% while maintaining accuracy, and in some cases by as much as 95\%~\citep{colledani2025machine,graham2019computerized}.  Multidimensional IRT (MIRT) extends this principle to model multiple correlated traits at once~\citep{chalmers2016generating}, offering a powerful mechanism for assessing comorbid conditions, though its application in language-based assessment remains unexplored. 

% Current Limitations and Our Contribution

While adaptive testing has gained significant traction in educational settings, its multidimensional applications remain relatively underexplored, particularly when leveraging open-ended language responses. 
Most existing approaches in NLP rely on unidimensional item response theory (IRT) focusing on single outcomes~\citep{lalor-etal-2016-building,varadarajan-etal-2024-alba}. 
To date, no prior work has effectively bridged this gap by integrating adaptive item selection with multitask learning for language-based, multidimensional mental health evaluation.

To our knowledge, this work is the first to tackle this challenge.  
 \textsc{MAQuA} combines multi-outcome modeling with multidimensional IRT (MIRT) to adaptively select open-ended questions for assessing multiple mental health conditions at once.
Our system uniquely models ten overlapping mental health constructs from targeted language data, learning to identify the most diagnostically informative questions while explicitly capturing their latent comorbid relationships.

\section{Background}
\label{sec:background}
With a growing need for scalable and nuanced approaches to mental health assessments, especially given that traditional clinical interviews and fixed-item scale assessments are limited by patient burden, clinician time, and difficulties in manually capturing multiple overlapping mental health conditions, Item Response Theory has emerged as an alternative measurement paradigm that enables adaptive assessments instead of traditional questionnaires and interviews.
\begin{algorithm}
\centering
\small
\caption{Adaptive Language-Based Assessment (ALBA)}
\begin{algorithmic}[1]
\STATE Initialize $\theta \leftarrow$ initial estimate of trait level
\STATE Initialize item prompt pool and empty response list $\texttt{responses} \leftarrow \emptyset$
\WHILE{$\texttt{responses}$, and stopping rule not met on $\theta$ }
    \STATE Select next item $p$ to maximize information at current $\theta$
    \STATE Present prompt $p$ and capture free-text response $t$
    \STATE Compute discrete response score $s$ = NLP\_score$(t)$
    \STATE Append $(p, s)$ to $\texttt{responses}$
    \STATE Update $\theta$ using IRT scoring method on $\texttt{responses}$
\ENDWHILE
\STATE Output final estimate $\theta$ 
\end{algorithmic}
\label{alg:alba}
\end{algorithm}

\textbf{Item Response Theory} (IRT) is a probabilistic, data-driven measurement framework that models the relationship between an individual’s latent trait score (such as depression severity) 
and their probability of specific item responses on a questionnaire~\cite{embretson2000item,hambleton1991fundamentals}. In single-dimensional IRT, this relationship is modeled with respect to one latent trait (usually denoted $\theta$), accounting for item-specific parameters such as difficulty and discrimination.
IRT enables precise ordering and calibration of items based on their informativeness in measuring the latent trait, supporting adaptive and individualized assessment. Adaptive language-based assessments~\citep{varadarajan-etal-2024-alba} were first introduced using single-dimensional IRT, summarized in Algorithm~\ref{alg:alba}.

\textbf{Exploratory Factor analysis (EFA)} is a statistical technique used to uncover the underlying structure of a set of observed variables by identifying clusters of variables that co-vary together, known as factors or latent constructs~\citep{cudeck2000exploratory}. Although it is very similar to Principal Component Analysis (PCA), it captures the multidimensional, often overlapping nature of latent variables by allowing factors to be correlated, unlike PCA which assumes uncorrelated components. This approach captures meaningful psychological constructs like depression and anxiety, supporting the development of sensitive, multidimensional assessments for accurate mental health assessments. 
EFA allows us to distill large and complex data such as responses to multiple questionnaires assessing various psychological conditions.
Each factor represents a distinct underlying construct that accounts for shared variance among the observed variables, providing insight into how different mental health symptoms or traits may be related at a deeper level. 

\textbf{Multidimensional item response theory (MIRT}) therefore relies on factor analysis (FA) to identify and model multiple latent traits underlying assessment items. FA reveals how items relate to different, but correlated psychological dimensions like depression or anxiety. These factor structures guide the MIRT model by linking items to specific traits, ensuring the model accurately captures the multidimensional nature of the data. This allows MIRT to estimate individuals’ scores across overlapping traits efficiently and precisely for multi-outcome assessment.
\footnote{Detailed background on multidimensional IRT is provided in the Table~\ref{tab:IRT}.}

\section{Dataset}
\label{sec:mulirt-data}

% \begin{figure}
%     \centering
% \includegraphics[width=\linewidth]{latent_figs/distribution_outcomes.png}
%     \caption{Distribution of standard questionnaire scales for all the participants in the collected dataset. AUDIT and DUDIT have a lot of shared symptomatology and the language-based questions were eliciting responses for substance usage, grouping alcohol and drugs in the same category. Further, not a lot of participants in the data collection were diagnosed with either alcohol use disorder or substance use disorder, so these were grouped for analysis by adding the two scores.}
%     \label{fig:dist_outcomes}
% \end{figure}

% \begin{figure}
%     \centering
%     \includegraphics[width=\linewidth]{IRT/mulirt/distribution_outcomes.png}
%     \caption{Distribution of standard questionnaire scales for all the participants in the collected dataset. AUDIT and DUDIT have a lot of shared symptomatology and the language-based questions were eliciting responses for substance usage, grouping alcohol and drugs in the same category. Further, not a lot of participants in the data collection were diagnosed with either alcohol use disorder or substance use disorder, so these were grouped for analysis by adding the two scores.}
%     \label{fig:dist_outcomes}
% \end{figure}

To explore these questions, we selected a unique set of material including language questions, diagnoses, and validated clinical scales. The data was collected in two phases: first, we pre-screened the participants to establish a diverse sampling pool including participants who stated to were diagnosed by a mental health professional. We focused on several common mental disorders, such as mood disorders (i.e., major depression, generalized anxiety disorder, bipolar disorder), autism spectrum disorder, attention-deficit/hyperactivity disorder, eating disorders, obsessive-compulsive disorder, post-traumatic stress disorder, and substance use disorders (i.e., alcohol and/or drug abuse). To enhance ecological validity, we intentionally included participants with comorbidities or co-occurring mental health illnesses, as they are the rule rather than the exception in clinical practice. In a second phase, we collected mental health data from these screened individuals using standardized rating scales, free-text narratives, and open-response questions targeting nine common mental disorders. 
\begin{figure}[!ht]
    \centering \includegraphics[width=.75\linewidth]{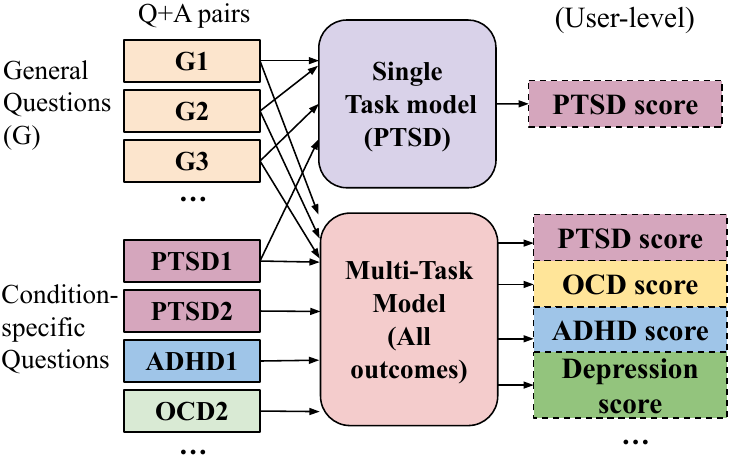}
    \caption{Single-task models are set up to predict a mental health condition score based on language responses to the general questions as well as the condition-specific questions. Multi-task models have been set up to take in all the language responses and predict all the mental health scores simultaneously.}
    \label{fig:method-diagram}
\end{figure}
We recruited a small set of 515 participants (297 female, 182 male, 35 non-binary, 1 who preferred not to categorize their gender identity) who were diagnosed with any of nine mental health conditions (Anxiety Disorders (AD), Bipolar Disorder (BD), Depression (D), Attention Deficit Hyperactivity Disorder (ADHD), Post-Traumatic Stress Disorder (PTSD), Obsessive-Compulsive Disorder (OCD), Eating Disorders (ED), Addiction and Substance Abuse (A), Autism (AU)). The age of the sample ranged from 18 to 78 (\textit{mean = 38.9}, \textit{SD = 12.3}) We screened participants for having one (or more) ongoing mental disorders (e.g., Np = 515, $\sim$50 of each disorder). The distribution of all the \textbf{diagnoses }across the participants is shown in Figure~\ref{fig:dist_diagnoses}. The participants first took a screening questionnaire for diagnoses and treatment for the ten mental health conditions to qualify for eligibility to participate in the survey. Here, 186 individuals reported receiving no treatment at all, while 329 reported receiving at least one form of treatment (i.e., medication and/ or psychotherapy). 

All participants then took ten rating scale questionnaires along with \textbf{language-based questions} related to all the mental health conditions considered.
A total of 48 language-response questions were developed based on DSM-5 criteria to capture key symptoms, frequency, and onset, and these items were reviewed by clinical psychologists to ensure clarity; 42 questions required descriptive-word responses and 5 essay responses. %check rated: yes or no are we including? there are 4
The dataset has two kinds of language questions posed to the participants: General (\textbf{GenQ}) and Condition-Specific (\textbf{CondQ}). There are 10 General Questions related to mental health in general, and all the other 38 questions are Condition-specific, mapping to the one of the questionnaire scales collected that we describe below.

 Ten \textbf{validated clinical scales} were also administered: 
the PHQ-9 (~\citet{kroenke2001phq}; 9 items, 4-point Likert) for depression, GAD-7 (\citet{spitzer2006brief}; 7 items, 4-point Likert) for anxiety, MDQ (\citet{miller2004sensitivity}; 14 binary items plus one 4-point Likert item) for bipolar disorder, RAADS-14 (\citet{eriksson2013raads}; 14 items, 4-point Likert) for autism, ASRS Part A (\citet{adler2006validity}; 6 items, 5-point Likert) for ADHD, NSESSS-PTSD (\citet{lebeau2014dimensional}; 9 items, 5-point 
Likert plus an open-text trauma description) for PTSD, BOCS (\citet{goodman1989yale}; 15 items, 3-point Likert plus an open-response categorization) for obsessive-compulsive symptoms, EDE-QS (\citet{fairburn2008eating}; 12 items, 4-point Likert) for eating disorders, and two substance use instruments: the AUDIT (\citet{allen1997review}; 8 items, 5-point Likert plus 2 items, 3-point Likert) for alcohol misuse and the DUDIT (\citet{berman2007dudit}; 9 items, 5-point Likert plus 2 items, 3-point Likert) for drug abuse.
Post-screening, we invited the selected participants to participate in the main study, with recruitment conducted via Prolific. The language questions, some eliciting descriptive words and some open-ended essay-like responses, are listed in Table~\ref{tab:questionnaire}. They were asked in random order to eliminate any systematic priming effects.

 \section{Methods}
We begin by exploring robustness of multidimensional models to estimate mental health trait scores from language responses. Next, we detail our application of factor analysis to uncover the underlying trait structure and map items to their respective dimensions. 
Finally, we describe the adaptive testing approach which leverages this factor structure within a multidimensional IRT framework to guide item selection, response scoring, and iterative trait estimation for efficient and precise multidimensional mental health assessment (See Algorithm~\ref{alg:mirt_alba_discretized}).

% \usepackage{graphicx}
% \usepackage[normalem]{ulem}

% Advancements in Item Response Theory~\citep{chalmers2016generating} have enabled modeling and adaptive testing for multidimensional factors, often in dichotomous and occasionally in polytomous response formats. Similar to \S\ref{chapter:alirt}, the conversion of language responses into polytomous scores, aka polytomization --  involves training regression models to score on each question for each outcome that it pertains to (for example, ``Have you been losing interest in everyday activities?'' is a question that pertains to a depression severity score) and then polytomizing based on the scores. However, the difference here is using multi-task objectives for predicting multiple outcome scores simultaneously based on the actual diagnoses reported by the participants, which we previously did not have access to. 

\subsection{Multi-outcome Modeling}
Given the multitude of psychological dimensions, associated language data, and the occurrence of comorbid diagnoses,
linguistic expressions intended to capture one dimension may also provide valuable information about others. 
To investigate this, we frame the language-based modeling of psychological scores in two configurations: single-task and multi-task. 
Using a stratified sampling approach based on depression outcomes (PHQ scores) over the users, we generate 9 folds for evaluation, 5 as train set, 3 as development set and 1 test set. Each fold contains the responses of around 58 users each. Language representations are derived from all text responses utilizing an encoder model. We experimented with popular models as shown in Table~\ref{tab:encoder}, and for the rest of the experiments, we used the \texttt{nomic-embed-text-v1.5} model~\cite{nussbaum2024nomic}, which are subsequently reduced to 16 dimensions through Matryoshka embeddings~\cite{kusupati2022matryoshka}. We then train linear regression models to predict standard mental health questionnaire scores across ten dimensions (including PHQ for depression, GAD for anxiety etc. as explained in \S\ref{sec:mulirt-data}). The single-task models predict each mental health dimension independently, whereas the multi-task models simultaneously predict all ten scores (see Figure~\ref{fig:method-diagram}). 
\noindent For model optimization, we performed hyperparameter tuning over ranges of learning rates, weight decay values, and used the Adam optimizer. We also compared output scaling methods for the regression, finding that min-max scaling consistently outperformed z-score normalization across all settings. Experiments were conducted using a single NVIDIA A6000 GPU. Further, we also prompt \texttt{llama3.2-1B} model~\cite{grattafiori2024llama} to compare the capacilities of zero-shot models in estimating the mental health condition scores (\ref{tab:llama32_prompt}). We also explore several model variants to analyze different aspects of the language data.
\paragraph{Comparison of Embedding models}
We compare various encoder models for multi-outcome modeling, the results are shown in Table~\ref{tab:encoder}. We find that Matryoshka reduction on the number of dimensions helps with modeling the less represented conditions (across all the conditions other than Depression and Anxiety).
%to create  and IRT training folds, as opposed to stratifying based on the rating scale outcome for the folds in \S\ref{chapter:alirt}. 
%Due to data scarcity, similar to previous work, we train the models in a cross-validation setting and analyze the combined predictions for polytomization.
\begin{algorithm}
\small
\caption{\textsc{\textsc{MAQuA}:} Multi-Adaptive Question-Asking}
\begin{algorithmic}[1]
% Step 1 - Modeling and Discretization
\STATE \textbf{Multi-outcome Modeling:} Multi-outcome regression models to capture mental health scores from language
\STATE Apply threshold-based discretization to transform continuous or modeled scores into discrete item-level responses suitable for factor analysis
\STATE \textbf{Factor Analysis:} Using discretized response data \textbf{X}, estimate factor loading matrix \(\mathbf{\Lambda}\) and latent factor scores \(\mathbf{f}_i\) for all individuals $i$
\STATE Determine number of factors $m$, factor correlations, and item-to-factor structure from \(\mathbf{\Lambda}\)

% Step 2 - Multidimensional IRT Model Construction
\STATE \textbf{Multidimensional IRT-based Adaptive Question Asking:}  
\STATE \hskip1em Initialize MIRT parameters $\{a_j, b_j, \mathbf{w}_j\}_{j=1}^p$ for each item $j$ based on $\mathbf{\Lambda}$ and factor structure
\STATE \hskip1em Set initial latent trait estimates $\boldsymbol{\theta}_i^{(0)}$ for each individual $i$

% Step 3 - Adaptive Testing Loop
\STATE Initialize item prompt pool and $\texttt{responses} \leftarrow \emptyset$
\WHILE{stopping criteria not met on $\boldsymbol{\theta}_i^{(t)}$ and pool not empty}
    \STATE For each candidate item $p$, compute Fisher information matrix $\mathcal{I}_p(\boldsymbol{\theta}_i^{(t)})$
    \STATE Select next item $p^*$ maximizing $\det(\mathcal{I}_{p^*}(\boldsymbol{\theta}_i^{(t)}))$ over all candidates (D-optimality)
    \STATE Present prompt $p$ to individual and capture text response $t$
    \STATE Compute multi-outcome discrete response score $\mathbf{s} = \text{NLP\_discretize\_score}(t)$ aligned with MIRT model item format
    \STATE Append $(p, \mathbf{s})$ to $\texttt{responses}$
    \STATE Update $\boldsymbol{\theta}_i^{(t+1)}$ using  maximum likelihood on $\texttt{responses}$
    \STATE $t \leftarrow t+1$
\ENDWHILE

\STATE \textbf{Output:} Final multidimensional trait estimates $\boldsymbol{\theta}_i^{(t)}$ for individual $i$

\end{algorithmic}
\label{alg:mirt_alba_discretized}
\end{algorithm}

\begin{table*}[!ht]
\resizebox{\textwidth}{!}{%
\begin{tabular}{ll|cccccccccc|c}
\toprule
 &  & \multicolumn{1}{l}{} & \multicolumn{10}{c}{\textbf{Pearson Correlation with Validated Clinical Scales}}\\ 
 %\cline{3-13}
\makecell{\textbf{Model}}& \makecell{\textbf{Aggr.}\\\textbf{type}}  & \makecell{\textbf{Depre-}\\\textbf{ssion}} & \multicolumn{1}{l}{\textbf{Anxiety}} & \multicolumn{1}{l}{\textbf{Bipolar}} & \multicolumn{1}{l}{\textbf{Autism}} & \makecell{\textbf{Drug}\\\textbf{use}} & \multicolumn{1}{l}{\textbf{OCD}} & \multicolumn{1}{l}{\textbf{ADHD}} & \multicolumn{1}{l}{\textbf{PTSD}} & \multicolumn{1}{l}{\textbf{Eating}} & \makecell{\textbf{Alcohol}\\\textbf{use}} & \multicolumn{1}{l}{\textbf{Avg.}} \\
%&  &\multicolumn{}& \multicolumn{1}{l}{\textbf{Anxiety}} & \multicolumn{1}{l}{\textbf{Bipolar}} & \multicolumn{1}{l}{\textbf{Autism}} & \makecell{\textbf{Drug}\\\textbf{use}} & \multicolumn{1}{l}{\textbf{OCD}} & \multicolumn{1}{l}{\textbf{ADHD}} & \multicolumn{1}{l}{\textbf{PTSD}} & \multicolumn{1}{l}{\textbf{Eating}} & \makecell{\textbf{Alcohol}\\\textbf{use}} & \multicolumn{1}{l}{\textbf{Avg.}} \\
\midrule
\multicolumn{2}{c}{LLaMA-3.2 (zero-shot)} &.179& .209&-.023 & .015 & .013&-.026&.019 &.049 &.007 & .087 & .052 \\
\midrule
Single Task & Input %& 10
& .763& .699& .398& .399& .351& .499& .424& .493& .409& .412& .485\\
Single Task & Output %& 48 
& .775& .703& .372& .425& .304& .569& .497& .468& .330& .355& .479\\
%\midrule
Multi Task & Input %& 1 
&\textbf{ .784}& \textbf{.722}& \textbf{.446}& \textbf{.449}& \textbf{.419}&\textbf{ .570}& \textbf{.560}&\textbf{ .532}& \textbf{.468}& \textbf{.478}& \textbf{.543}\\
Multi Task & Output %& 1 
& .433& .443& .401& .307& .394& .408& .380& .366& .387& .411& .389\\
\midrule
 \multicolumn{3}{l}{} & \multicolumn{10}{c}{\textbf{Pointwise Biserial Correlation with Diagnoses}}\\
 \multicolumn{2}{l}{} & \makecell{\textbf{Depre-}\\\textbf{ssion}} & \multicolumn{1}{l}{\textbf{Anxiety}} & \multicolumn{1}{l}{\textbf{Bipolar}} & \multicolumn{1}{l}{\textbf{Autism}} & \makecell{\textbf{Substance}\\\textbf{use}} & \multicolumn{1}{l}{\textbf{OCD}} & \multicolumn{1}{l}{\textbf{ADHD}} & \multicolumn{1}{l}{\textbf{PTSD}} & \multicolumn{1}{l}{\textbf{Eating}} & \makecell{-} & \multicolumn{1}{l}{\textbf{Avg.}} \\
 %\cline{3-13}
%\makecell{\textbf{Model}\\\textbf{name}}& \makecell{\textbf{Aggr.}\\\textbf{type}}  & \makecell{\textbf{Depre-}\\\textbf{ssion}} & \multicolumn{1}{l}{\textbf{Anxiety}} & \multicolumn{1}{l}{\textbf{Bipolar}} & \multicolumn{1}{l}{\textbf{Autism}} & \makecell{\textbf{Drug}\\\textbf{use}} & \multicolumn{1}{l}{\textbf{OCD}} & \multicolumn{1}{l}{\textbf{ADHD}} & \multicolumn{1}{l}{\textbf{PTSD}} & \multicolumn{1}{l}{\textbf{Eating}} & \makecell{\textbf{Alcohol}\\\textbf{use}} & \multicolumn{1}{l}{\textbf{Avg.}} \\
%&  &\multicolumn{}& \multicolumn{1}{l}{\textbf{Anxiety}} & \multicolumn{1}{l}{\textbf{Bipolar}} & \multicolumn{1}{l}{\textbf{Autism}} & \makecell{\textbf{Drug}\\\textbf{use}} & \multicolumn{1}{l}{\textbf{OCD}} & \multicolumn{1}{l}{\textbf{ADHD}} & \multicolumn{1}{l}{\textbf{PTSD}} & \multicolumn{1}{l}{\textbf{Eating}} & \makecell{\textbf{Alcohol}\\\textbf{use}} & \multicolumn{1}{l}{\textbf{Avg.}} \\
\midrule
\multicolumn{2}{c}{Clinical Scales (upper)} &.404&.423&.440&.454&.097 &.133&.073 &.172&.080 & -&.253\\
\multicolumn{2}{c}{LLaMA-3.2 (zero-shot)}& .032& .104& .036& 034& .065& -.005& .029& .071& .069& -& .048\\
\midrule
Single Task & Input %& 10
& .346& .333& .220& .036& .165& .136& .081& .160& .078& .& .173\\
Single Task & Output %& 48 
& \textbf{.389}& .393& .135& .170& .123& .146& .179& \textbf{.269}& .037& .& .205\\
%\midrule
Multi Task & Input %& 1
& \textbf{.388}& \textbf{.428}& \textbf{.244}& \textbf{.218}& \textbf{.269}& \textbf{.173}& \textbf{.195}& .249& \textbf.108& -& \textbf{.252}\\
Multi Task & Output %& 1 
& .379& .415& .228& .139& .190& .127& .182& .239& \textbf{.149}& -& .227\\
\bottomrule
\end{tabular}%
}
\caption{Comparison of Aggregation types and Task Formulation to predict multiple psychological scores from language. \textbf{Bold} indicates significance against the second best in the column. %\vasudha{add experiments for: comparison against ALBA (single dim), more embedding representations (roberta, all dim vs reduced), estimating and ordering capabilities of LLMs (maybe 2-3)}
}
\label{tab:single_v_multi}
\end{table*}

\begin{table*}[h]
  \centering
  \begin{minipage}{0.55\textwidth}
    \centering
\resizebox{\columnwidth}{!}{
\begin{tabular}{lcccc|ccc}
\toprule
&\multicolumn{4}{c}{\textbf{with Q (Lang)}}&\multicolumn{2}{c}
{\textbf{Ablation}}\\

 & \makecell{ \textbf{GenQ} } &   \textbf{CondQ} & \makecell{\textbf{CondQ}\\\textbf{+ GenQ}} &\textbf{All Q}& \makecell{\textbf{With}\\\textbf{Q (ID)}}& \textbf{No Q}\\
 \midrule
 
Depression  & .782&  .785&.785& .784&.795&.792\\
Anxiety & .721& .724& .723& .722 & .725 &.724\\
Bipolar & .440& .440& .432& \textbf{.446}&.394&.390\\
Autism & .450& .451& .444& .449& .423& .429\\
Drug use & .423& \textbf{.431}& .423& .419& .324& .272\\
OCD & .566& \textbf{.574}& .570& .570& .561& .568\\
ADHD & .558& \textbf{.560}& .554& \textbf{.560}& .532& .532\\
PTSD & .534& .538& .536& .532& .490& .493\\
Eating & .458& .463& .459& \textbf{.468}& .416& .335\\
Alcohol use & .469& .474& .465& \textbf{.478}& .374& .359\\
\midrule
Average & .540& \textbf{.544}& .539& \textbf{.543}& .503& .489\\
\bottomrule
\end{tabular}}
\caption{%What kinds of question makes a difference in the modeling? We find that none of the disorders benefit solely from general questions. What is interesting is that using all the questions (including other condition-specific) does improve the prediction of anxiety, bipolar and eating disorders. We also explore the effect of including the question -- it improved when language of question is included.
Comparison of inclusion of various types of questions in Multi-outcome Modeling, and the effect of ablation of question embeddings from the Q-A input representations.}
\label{tab:question_ablation}
  \end{minipage}%
  \hfill
  \begin{minipage}{0.43\textwidth}
\small
\centering
\resizebox{\columnwidth}{!}{
\begin{tabular}{lccc}
\toprule
\makecell{\textbf{Mental Health }\\\textbf{Condition}}& \makecell{\textbf{Factor 1}\\\textbf{Loading}} & \makecell{\textbf{Factor 2 }\\\textbf{Loading}} & \makecell{\textbf{Dominant}\\\textbf{Factor}}\\
\midrule
Depression             & .908& .210& 1 \\
Anxiety                & .953& .198& 1\\
Bipolar                & .779& .546& 1,2\\
Autism                 & .861& .063& 1\\
Substance use               & .305& .870& 2\\
OCD                    & .945& .240& 1\\
ADHD                   & .918& .274& 1\\
PTSD                   & .716& .430& 1,2\\
Eating disorder        & .091& .928& 2\\
Alcohol use            & .672& .418& 1,2\\
\bottomrule
\end{tabular}}
\caption{Exploratory factor analysis results for mental health conditions. The dominant factor indicates which factor has the highest loading for each condition. Bipolar, PTSD, and alcohol use disorder are modeled as cross-loadings.}
\label{tab:mulirt_fac_analysis}
  \end{minipage}
\end{table*}

\paragraph{1. Aggregation type}
%Basically two ways to aggregate -- should we aggreagete the language itself or should we aggregate the final predictions or model outputs? 
%Input aggregating -- averaging the embeddings of the input language responses and predicting a user-level score.
%Output aggregating -- each language response is independently modeled -- there is a model for each question, predicting a score pertaining to the mental health condition score of that question. Then the questions belonging to the same condition (includign general) are aggreagated to produce a predicted condition score. 
To aggregate multiple language responses from each participant, we consider two main approaches. The first, input aggregation, involves averaging the embeddings of all input responses for a user and then using this combined representation to predict an overall mental health score at the user level. The second approach, output aggregation, treats each language response separately: a model predicts a mental health score for each response, and then these scores are combined by averaging or another method for all questions related to the same condition to produce a final predicted score for that condition. This allows us to compare whether it is more effective to aggregate at the language level or the level of model predictions. 

\paragraph{2. Question Information} We explore the role of the language of question wording on the modeling of mental health outcomes. Unlike conventional language-based assessments that rely on ecological data such as social media posts, our setting is distinct because the language analyzed is generated as direct responses to specific prompts. To understand whether the phrasing of the questions themselves affects the models, we conduct an ablation study that incorporates the question ID as an input feature. This allows us to disentangle whether it is the unique identity of the question, rather than its linguistic content, that primarily drives the modeling performance.

 We report the Pearson correlation of the predicted scores against the validated clinical scales they were originally trained on. Further, we also report the Pointwise Biserial correaltion of the predicted scores against each of the nine diagnoses collected (binary-valued). The results are shown in Tables~\ref{tab:single_v_multi},~\ref{tab:question_ablation}.
The best model determined was then used to train across 9 folds with 4 folds as  the regression task train set, 4 folds as MIRT train set and 1 fold for MIRT test set, and \textsc{MAQuA} was run on a 9-fold cross validation, reporting the aggregated scores across all the test sets. This design choice was made to ensure enough training data for both the multi-outcome modeling as well as the MIRT modeling.

\begin{table*}[!h]
\centering
\resizebox{\linewidth}{!}{
\begin{tabular}{p{0.25\linewidth} c c c c c c c c c c c c}
\toprule
\textbf{Model} & \makecell{\textbf{Depre-}\\\textbf{ssion}} & \multicolumn{1}{l}{\textbf{Anxiety}} & \multicolumn{1}{l}{\textbf{Bipolar}} & \multicolumn{1}{l}{\textbf{Autism}} & \makecell{\textbf{Substance}\\\textbf{use}} & \multicolumn{1}{l}{\textbf{OCD}} & \multicolumn{1}{l}{\textbf{ADHD}} & \multicolumn{1}{l}{\textbf{PTSD}} & \multicolumn{1}{l}{\textbf{Eating}} & \makecell{\textbf{Alcohol}\\\textbf{use}} & \multicolumn{1}{l}{\textbf{Avg.}}  & \textbf{Number of dim} \\
\midrule
nomic-embed-text-v1 & .784& .722& .446& .449& .419& .570& .560& .532& .468& .478& .543 & 16\\
\midrule
nomic-embed-text-v1  & .763& .688& .442& .487& .239& .426& .403& .265& .270& .232& .421& 768\\
mxbai-embed-large-v1 & .758& .668& .480& .484& .246& .434& .406& .240& .298& .150& .417& 1024\\
roberta-base & .733& .658& .394& .428& .148& .441& .298& .234& .252& .231& .382& 768\\
roberta-large & .780& .675& .385& .449& .210& .429& .418& .269& .274& .225& .411& 1024\\
\bottomrule
\end{tabular}
}
\caption{Comparison of popular embeddings~\cite{nussbaum2024nomic,emb2024mxbai,liu2019roberta} derived from encoder models, for multitask, input-aggregated setting. While any of these models could be used with our framework, we find Matryoshka reduction to be particularly helpful in modeling in low-resource settings.}
\label{tab:encoder}
\end{table*}

% \begin{table}
% \centering
% \small
% \begin{tabular}{lccc }
% \toprule
%  &\textbf{ With Q (lang)}& \textbf{With Q (num)}& \textbf{No Q}\\
%  \midrule
% Depression & .784& \textbf{.795}& .792\\
% Anxiety & .722& .725& .724\\
% Bipolar & \textbf{.446}& .394& .390\\
% Autism & \textbf{.449}& .423& .429\\
% Drug use & \textbf{.419}& .324& .272\\
% OCD & .570& .561& .568\\
% ADHD & \textbf{.560}& .532& .532\\
% PTSD & \textbf{.532}& .490& .493\\
% Eating & \textbf{.468}& .416& .335\\
% Alcohol use & \textbf{.478}& .374& .359\\
% \midrule
% Average & \textbf{.543}& .503& .489\\
% \bottomrule
% \end{tabular}

% \caption{We also explore the effect of including the question -- gives signals to distinguish condition-specificty of question }
% \end{table}

\subsection{Factor Analysis}
Multidimensional adaptive question-asking algorithm requires setting up a factor model, which is defined through factor analysis to determine the factor structure, that takes in all the questions.
We run exploratory factor analysis to determine the optimal factor structure for this dataset for the predicted user-level scores. The factor loadings are reported in Table~\ref{tab:mulirt_fac_analysis} and Figure~\ref{fig:factor_words}. Then we run aggregate the multi-outcome model predictions at a question level, across all users, to analyze how each question loads on to the factors by by applying the factor model at the question-level. This yields a question-level loading which is used to define the MIRT model (Table~\ref{tab:question-level}).
%and (b) predicted question-level scores.
% \begin{table}[!ht]
% \small
% \centering
% \begin{tabular}{lccc}
% \toprule
% \makecell{\textbf{Mental Health }\\\textbf{Condition}}& \makecell{\textbf{Factor 1}\\\textbf{Loading}} & \makecell{\textbf{Factor 2 }\\\textbf{Loading}} & \makecell{\textbf{Dominant}\\\textbf{Factor}}\\
% \midrule
% Depression             & .908& .210& 1 \\
% Anxiety                & .953& .198& 1\\
% Bipolar                & .779& .546& 1,2\\
% Autism                 & .861& .063& 1\\
% Substance use               & .305& .870& 2\\
% OCD                    & .945& .240& 1\\
% ADHD                   & .918& .274& 1\\
% PTSD                   & .716& .430& 1,2\\
% Eating disorder        & .091& .928& 2\\
% Alcohol use            & .672& .418& 1,2\\
% \bottomrule
% \end{tabular}
% \caption{Exploratory factor analysis results for mental health conditions. The dominant factor indicates which factor has the highest loading for each condition. Bipolar, PTSD, and alcohol use disorder are modeled as cross-loadings.}
% \label{tab:mulirt_fac_analysis}
% \end{table}
\begin{table}[!ht]
\resizebox{\linewidth}{!}{
\begin{tabular}{cp{5cm}p{5cm}}
\toprule
\textbf{Factor} & \textbf{Question IDs} & \textbf{Condition/Symptom} \\
\midrule
F1 & OMD1, OMD2, OMD3, OMD4, OMD5, OMD6, A1, A3, A4, BD2, BD3, ASD2, ASD3, ASD4, ASD5, ASD6, OCD1, OCD2, OCD3, ADHD1, ADHD2, PTSD1, PTSD2, PTSD4, ED1, ED2, ED3, ED4, ED5, ED6 & Depression (OMD), Anxiety (A), Bipolar Disorder (BD), Autism Spectrum Disorder (ASD), Obsessive-Compulsive Disorder (OCD), Attention Deficit Hyperactivity Disorder (ADHD), Post-Traumatic Stress Disorder (PTSD), Eating Disorders (ED) \\
\midrule
F2 & BD2, BD3, SUB1, SUB2, SUB3, SUB4, SUB5, SUB6, OCD1, OCD2, OCD3, ED1, ED2, ED3, ED4, ED5, ED6 & Bipolar Disorder (BD), Substance Use (SUB), Obsessive-Compulsive Disorder (OCD), Eating Disorders (ED) \\
\bottomrule
\end{tabular}}
\caption{Question-level mapping to the two significant factors that was further used to specify the MIRT model (see Table~\ref{app:questionnaire} for legend).}
\label{tab:question-level}
\end{table}
We find that two significant factors emerged based on parallel analysis~\cite{horn1965rationale}. The first factor is characterized by strong loadings from measures of depression, anxiety, PTSD, ADHD, and autism, suggesting that it reflects a broad \textit{internalizing} or emotional distress dimension. The second factor is dominated by high loadings for drug and alcohol use, pointing toward a substance use or \textit{externalizing} dimension. 
Conditions such as bipolar disorder, PTSD, and alcohol use disorders exhibit notable loadings on both factors  (within 20\% of each other), indicating that they share features with both internalizing and externalizing constructs.

%We define a 9-factor  (9 standard questionnaire outcomes) model with each factor mapping to each condition. The 18 general items are defined to be a part of all the nine factors. 

\subsection{Adaptive Testing}
We use the \texttt{mirtCAT}\footnote{\url{https://CRAN.R-project.org/package=mirtCAT}}, a computerized adaptive testing framework based on \texttt{mirt} to implement adaptive testing. 
The adaptive testing is done by choosing the most informative question at each turn, determined using D-optimality, a heuristic determined to be optimal for multidimensional IRT in \cite{chalmers2016generating}. It maximises the determinant of the Fisher information matrix at each turn of the questions, thus maximizing information gain across all the underlying dimensions. 

\paragraph{Discretization}
Since IRT models operate on discrete, ordered responses, continuous scores obtained from regression cannot be directly used as inputs. Instead, these continuous predictions must be discretized into \( k \) ordinal levels to allow the model to estimate category response functions. This discretization enables the IRT model to capture how each level probabilistically relates to underlying latent dimensions, following standard practices in multidimensional IRT modeling \cite{reckase200618}. 
We select the maximum number of discrete levels that ensures sufficient representation across the training data, following the heuristics in prior work on discretizing language-based scores for IRT~\cite{varadarajan-etal-2024-alba}. We discretize scores into four ordinal levels based on the quartile thresholds of the model outputs. Using greater granularity leads to data sparsity, as certain levels would be underrepresented, reducing model stability and interpretability.

\begin{figure*}[!h]
    \centering
    \includegraphics[width=.8\linewidth]{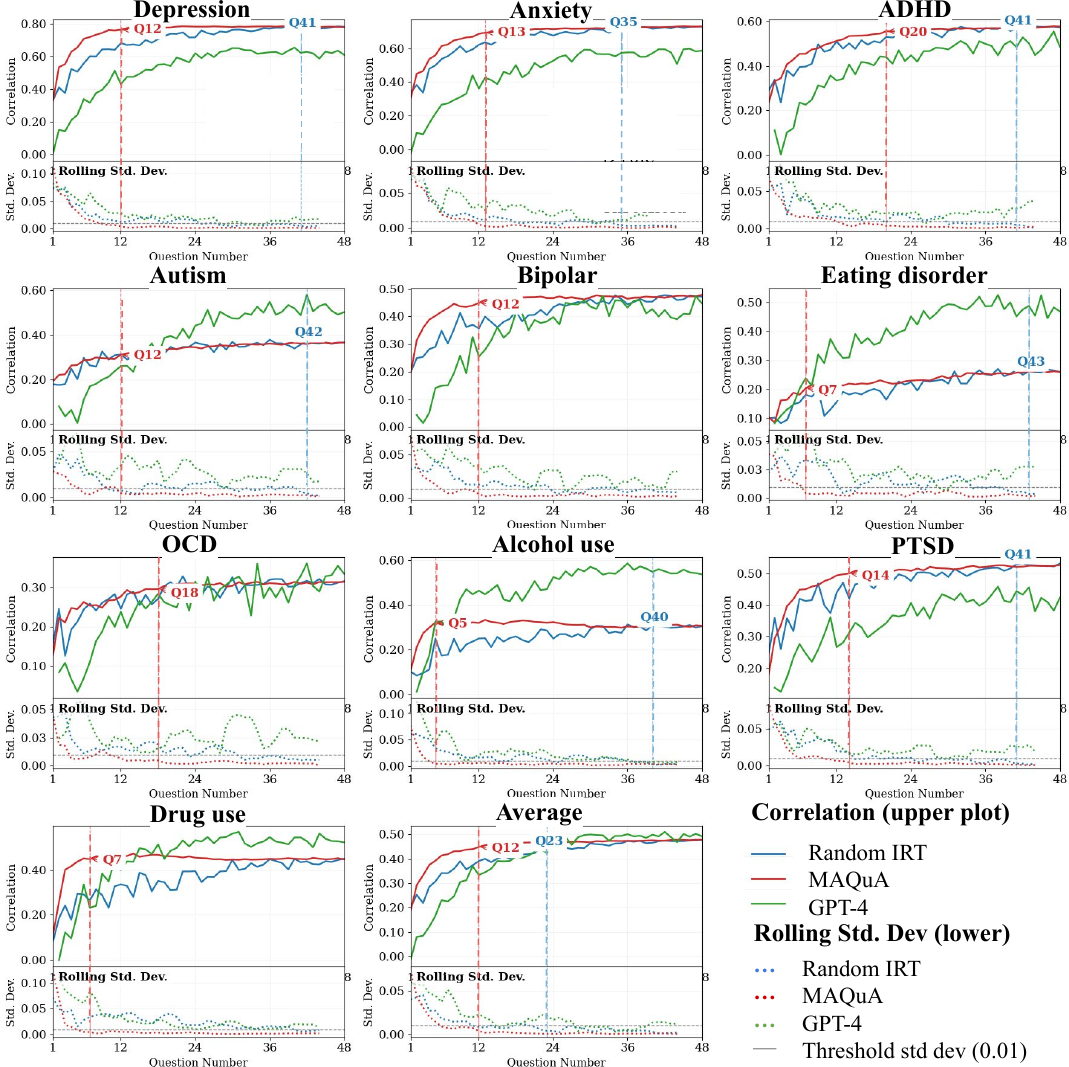}
    \caption{Pearson correlations of \textsc{MAQuA}-estimated scores over the number of questions asked along with their rolling standard deviation of the correlations. The vertical line shows the stability of the estimation based on a threshold for the standard deviation. Our adaptive method consistently stabilizes in at most 50\% the number of questions as random. While GPT-4 shows promise in its estimation capabilities of factor 2 (externalizing disorders) - substance use, alcohol use, eating disorder. For more details please see \S\ref{app:stabilization_points} .}
    \label{fig:stabilization}
\end{figure*}

We compare random question ordering to adaptive sequencing using \textsc{MAQuA}, and further benchmark these against GPT-4, one of the most popular LLMs previously studied for its mental health estimation capabilities and ability to handle long contexts~\citep{moell2024comparing,gpt4openai2023, ganesan2024explaining}. At each step, we report the Pearson correlations between estimated scores and validated clinical scales, summarized in Figure~\ref{fig:stabilization} and Table~\ref{app:stabilization_points}. The prompt can be found in Table~\ref{tab:gpt4_prompt}.

% We compare random ordering of the questions to that of adaptive ordering using \textsc{MAQuA}, as well as against the adaptive question-asking and estimation capabilities of one of the most popular LLMs that can handle long contexts -  gpt4 . We report the Pearson correlations of the estimated scores at each step against the validated clinical scales in Table~\ref{tab:condition_scores}. 
% We compare random question ordering to adaptive sequencing using \textsc{MAQuA}, as well as to the adaptive question selection and estimation abilities of GPT-4, a widely used LLM capable of handling long contexts. At each step, we compute the Pearson correlation between the estimated scores and established clinical scales; these results are presented

Further, we also compare the ordering and estimation capabilities of \texttt{gpt4} to showcase the efficacy of IRT in adaptive question-asking.
We also calculate the stability of standard deviation of the estimates at each turn of question-asking, setting the threshold at 0.01 to determine the point at which the estimated mental health scores from \textsc{MAQuA} for each dimension mostly stabilizes and approaches convergence. This metric can also help determine early stopping when deploying \textsc{MAQuA} (Figure~\ref{fig:stabilization}).

% To determine the best models for adaptive testing, we use correlations with the standard questionnaire scores as well as the diagnoses (which the regression models were not trained to predict
\section{Results}
\paragraph{Multi-outcome Modeling}
The results in Table~\ref{tab:single_v_multi} compare the predictive performance of single-task and multi-task models, 
as well as different aggregation strategies, across ten psychological dimensions. Multi-task models generally achieved higher correlations than single-task models when using input-level aggregation, with a strong performance across all the ten dimensions averaging at a correlation of 0.543 against the validated clinical scales and 0.252 against diagnoses. 

In the case of single-task models, output-level aggregation performed better over input-level aggregation, presumably due to missing data points due to skipped responses for the less common conditions. Surprisingly, multi-task model with output aggregation performs the worst across all the dimensions; this could be because of all the individual questions not being relevant to most of the dimensions at the same time, forcing spurious correlations to be meaningful signals. 
In general, all models performed higher on internalizing (depression, anxiety, OCD) factors as opposed to externalizing (substance use). Most interestingly, we found that with respect to diagnoses, the models performed better than some of the validated clinical scales. This could be caused by over-representation of certain diagnoses in the dataset, or (outdated / mis)diagnoses.

Table~\ref{tab:question_ablation} shows that including all the questions, General (GenQ) as well as Condition-specific (CondQ), alongside non-Condition-specific questions (i.e., the condition-specific questions that are related to other conditions and not the considered condition) performs as well as using just Condition-specific questions. In particular, drug use, OCD and ADHD are best captured with Condition-specific questions alone. Further, on performing ablation with the language of the question, we find that the questions, including just the question ID, indeed add context to modeling multiple outcomes at once.
% Multi-task models with input-level aggregation performed notably lower on most constructs, except for drug use, where a correlation of 0.500 was observed.
% When multi-task models used output-level aggregation, their predictive performance increased, with the highest correlations for anxiety (r = 0.462), bipolar (r = 0.347), and PTSD (r = 0.369), and an average correlation of 0.334. Overall, single-task models with input-level aggregation demonstrated superior performance for most psychological constructs, while multi-task models were more effective for specific outcomes and benefited from output-level aggregation.
% Therefore, for adaptive testing, we discretize predictions from the single-task output setting.

\begin{figure}[!ht]
    \centering
    \includegraphics[width=.85\linewidth]{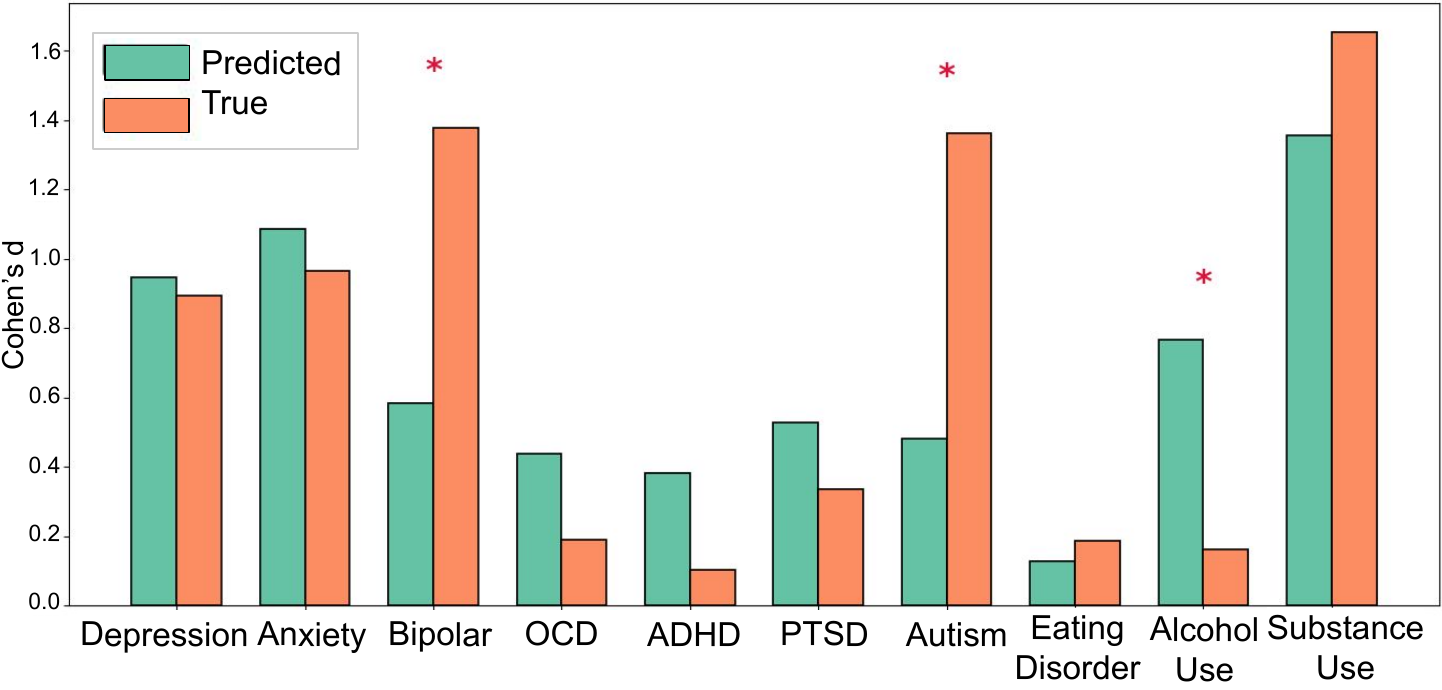}
    \caption{Cohen's d against the reported diagnoses for our best multitask model against the validated clinical scores (considered \textit{ground truth} in the modeling). \textcolor{red}{*} indicates one being significantly better correlated to diagnosis than the other.}
    \label{fig:cohens-d}
\end{figure}

\paragraph{Multi-outcome Model Prediction Structure}
We show the difference between how the ``\textit{ground truth}'' or validated clinical scores are related to diagnoses as compared to the predicted scores from the best performing multi-outcome model in Figure~\ref{fig:cohens-d}.  Among ten conditions, only three conditions show significant differences -- Bipolar, Autism and Alcohol use. While Bipolar and Autism are better captured by the clinical scores, the predicted scores for Alcohol Use actually outperform the clinically validated scores (AUDIT), showing that alcohol use might be better captured when the other conditions are taken into account as well. Moreover, diagnoses could be preliminary and \textit{wrong}~\cite{mendel2011confirmation}, since it could be a proxy for some other mental health condition. This could lead to spurious correlations that are better disambiguated with signals from a multitask model.

\paragraph{Adaptive Question-Asking}
Figure~\ref{fig:stabilization} shows the when the estimated scores from \textsc{MAQuA} are significantly better than their random counterpart, as well as GPT-4. After the first question, \textsc{MAQuA} seems to consistently jump to improve estimates across all dimensions, whereas in the case of random, the jumps are inconsistent. 
Based on the correlation curves presented in Figure~\ref{fig:stabilization}, all the three methods can lead to a plateau in score estimation across most of the dimensions, suggesting that modeling multiple conditions can lead to the model learning the shared semantics across the conditions, regardless of the question-asking strategy used. \textsc{MAQuA} doesn't always converge better: in the case of ADHD, Autism and OCD, random question-asking is just as good as \textsc{MAQuA} in terms of when convergence (small change in estimated scores in subsewuent turns) or maximum performance is reached. Despite that, \textsc{MAQuA} outperforms random ordered question-asking, especially for conditions like bipolar, alcohol and drug use, depression, anxiety and eating disorder, especially in the first few turns (2-12). This suggests that both the factors, \textit{internalizing} and \textit{externalizing}, are being prioritized almost equally when optimizing for information gain over multiple turns of question-asking, which indicates that the chosen model is quite effective in adaptive question-asking while optimizing across ten distinct conditions. 

In contrast, GPT-4’s estimation and question-asking behavior differ. Despite being prompted to assess all ten conditions simultaneously (rather than focusing on two factor scores), GPT-4 notably excels over the IRT-based methods in handling externalizing factors: specifically eating disorder, substance use, and alcohol use conditions. This advantage may stem from limitations inherent in the item response theory model, particularly regarding its representation of comorbidity effects within each data subset. Evidence supporting this includes the regression model’s performance, which approximates GPT-4’s correlation levels (Table~\ref{tab:single_v_multi}).
These findings suggest that a hybrid modeling approach could more effectively capture both internalizing and externalizing aspects.

To determine early stopping criteria and when a question-asking session could be potentially shortened, we also report the stabilization points in Figure~\ref{fig:stabilization} by marking the question numbers. Stabilization does not necessarily indicate peak performance, it indicates slower ascent after that point. The vertical lines indicate the point (n$^{th}$ question) where the rolling standard deviation drops below a threshold. As reported in Table ~\ref{tab:condition_scores}, employing the early stopping rule could lead to $50-85$\% reduction in the number of questions across all the mental health conditions being evaluated simultaneously. 
Figure~\ref{fig:stabilization} shows stabilization points for individual and averaged correlation across all the ten conditions. We find that on an average, \textsc{MAQuA} takes about $12$ questions to reach the stabilization point (threshold$=0.01$)  whereas random question-asking takes $24$ and GPT-4 doesn't converge at $48$ questions.
This offers significant potential to save time for both LLM-patient and clinician-patient interactions while reducing overall burden.

% Our research builds on recent progress in making mental health assessments more accurate, efficient, and realistic.
% %thanks hui! We don't usually have subsections for related work but I'll use this and work on it!
% %\subsection{Smarter and Faster Ways to Assess Mental Health}
% For years, mental health assessment has relied on rating scales. Recently, two major improvements have emerged.
% First, studies show that we get highly accurate results (with correlations over 0.8) when we ask standard mental health questions but let people answer in their own words instead of with numbers~\citep{kjell2022natural,varadarajan-etal-2024-alba,sikstrom2023precise}. This proves that language is a powerful source of information.
% Second, adaptive testing has made assessments much faster. These smart tests can replace hours-long clinical interviews with just a few tailored questions~\citep{gibbons2016computerized}. They are so effective that they have been used to predict suicide attempts in young people months in advance~\citep{king2021prospective}.
\section{Conclusion}

This work presents \textsc{MAQuA}, a novel adaptive, language-based framework that enables efficient, 
simultaneous assessment of multiple mental health dimensions by leveraging the strengths of modern language modeling along with multidimensional item response theory. Our empirical findings demonstrate that multi-task modeling with both shared and unique linguistic features significantly improves predictive accuracy across ten distinct mental health outcomes compared to single-task baselines. Moreover, by integrating adaptive question selection optimized for information gain across multiple dimensions, \textsc{MAQuA} substantially reduces the number of questions required to achieve stable diagnostic estimates, cutting patient burden by up to 85\% without compromising accuracy.

These results highlight the potential in LLM mental health agents for combining advanced language understanding with psychometrically grounded adaptive testing to overcome limitations of prior approaches that face challenges related to the inconsistency of LLM responses, and are not optimized to reduce patient burden. \textsc{MAQuA}’s effectiveness in modeling transdiagnostic symptom profiles marks an important step toward scalable, interactive, and clinically valid mental health assessment tools. Future research should extend and test this framework on real-time conversational settings, improving  generalizability in diverse clinical populations. %Through scalable and efficient modeling, \textsc{MAQuA} furthers the evolving AI-for-healthcare landscape by offering a paradigm for multidimensional mental health evaluation that aligns with real-world clinical needs.

\section*{Limitations}
This work has several important limitations. First, all participants provided responses in English and were primarily recruited from the UK, Sweden, and the US, which may restrict the applicability of our findings to other languages and cultural settings.
Additionally, ADHD is underrepresented in the dataset, limiting the reliability of conclusions related to this condition. 

We use a fixed set of questions to maintain control over the assessment and prevent large language models from hallucinating or generating harmful, misleading questions. Although this limits open-ended question selection, it is an important safeguard for participant safety and assessment reliability. Future work will explore flexible approaches with robust hallucination mitigation strategies.
While the multitask capabilities of large language models (LLMs) are critical, they were not explicitly explored in this study.
Specifically, the effects of reinforcement learning from human feedback (RLHF) or direct preference optimization (DPO) on question sequencing and preferences are unexplored and warrant further investigation.

Although our multi-outcome models for mental health assessment are not fully accurate for clinical diagnosis, we proceed with modeling downstream question-asking since it more closely mirrors how such models would be deployed in real-life settings. However, these models have not been tested in actual clinical environments and should not be used for diagnosis. They are instead intended as screening tools that may complement therapists and clinicians within their processes.

 \section*{Ethical Considerations}

Participants were compensated hourly for their time, ensuring fair recognition of their contribution and respecting the principle of voluntary and informed participation. All responses were anonymized to protect participant confidentiality, and the data were stored securely on protected servers to safeguard privacy and comply with data protection standards. The study received approval from the Institutional Review Board (IRB) [redacted], assuring that appropriate ethical oversight was in place throughout the research process.

\textsc{MAQuA} is designed to reduce the burden of mental health assessments and address diagnostic inconsistencies, thereby promoting ethical principles of minimizing participant fatigue and improving assessment accuracy. By combining multi-outcome modeling with multidimensional item response theory (MIRT), \textsc{MAQuA} selects the most informative questions adaptively, reducing assessment length by up to 85\% while maintaining psychometric validity, which supports participant well-being and respects their time.

\textsc{MAQuA} represents a methodological integration of advanced natural language processing and psychometric measurement theory, grounded in ethical commitments to scientific integrity, transparency, and reproducibility. The release of the accompanying questionnaire-driven dataset encourages further validation and responsible innovation, ensuring that AI tools in mental health are developed with accountability and sustained patient trust.

\bibliography{references, example_paper}
%\bibliographystyle{acl_natbib}

%%%%%%%%%%%%%%%%%%%%%%%%%%%%%%%%%%%%%%%%%%%%%%%%%%%%%%%%%%%%%%%%%%%%%%%%%%%%%%%
%%%%%%%%%%%%%%%%%%%%%%%%%%%%%%%%%%%%%%%%%%%%%%%%%%%%%%%%%%%%%%%%%%%%%%%%%%%%%%%
% APPENDIX
%%%%%%%%%%%%%%%%%%%%%%%%%%%%%%%%%%%%%%%%%%%%%%%%%%%%%%%%%%%%%%%%%%%%%%%%%%%%%%%
%%%%%%%%%%%%%%%%%%%%%%%%%%%%%%%%%%%%%%%%%%%%%%%%%%%%%%%%%%%%%%%%%%%%%%%%%%%%%%%
\newpage
\appendix
%\onecolumn
\renewcommand\thefigure{A.\arabic{figure}}    
\setcounter{figure}{0} 
\renewcommand\thetable{A.\arabic{table}}    
\setcounter{table}{0} 

\section{Details on Multidimensional IRT}
A detailed description of the modeling of MIRT is described in Table~\ref{tab:IRT}. Most of the terms and explanations were derived from ~\citet{chalmers2012mirt}.
\begin{table*}
\centering
\resizebox{\linewidth}{!}{
\begin{tabular}{ l l}
\toprule
\multicolumn{2}{c}{\textbf{Multidimensional IRT details}}\\
\midrule
Description &\makecell[l]{Multidimensional Item Response Theory (MIRT) extends classical unidimensional item response theory (IRT) \\to better capture complex psychological constructs by modeling multiple underlying dimensions of symptoms,\\ rather than a single overall trait. This is especially important for mental health assessments, since symptoms \\often span affective, cognitive, and physical domains that interact and overlap, making a nuanced,\\ multidimensional model necessary for accurately representing mental health scores.} \\
\midrule
Latent Trait Vector & \makecell[l]{
In multidimensional item response theory (MIRT), the latent trait vector is defined as:\\
\begin{math}
    \boldsymbol{\theta} = (\theta_1, \theta_2, \ldots, \theta_m),
\end{math}
\\
where each component \(\theta_k\) represents the individual's standing on the \(k\)-th latent dimension. This vector\\ characterizes the respondent's abilities or traits across multiple correlated or independent dimensions.\\
The latent traits \(\boldsymbol{\theta}\) are typically assumed to follow a multivariate normal distribution:\\
\begin{math}
\boldsymbol{\theta} \sim \mathcal{N}(\boldsymbol{\mu}, \boldsymbol{\Sigma}),
\end{math}
\\
where \(\boldsymbol{\mu}\) is the mean vector (often 0) and \(\boldsymbol{\Sigma}\) is the covariance matrix capturing correlations among the\\ latent traits.
This multivariate representation enables modeling the probability of a particular response \\to an item as a function of these multiple latent traits and item parameters in a probabilistic framework.
} \\
\midrule
Item Parameters & \makecell[l]{
Each item \(j\) in a multidimensional item response theory (MIRT) model is characterized by a discrimination vector:\\ 
\begin{math}
\boldsymbol{a}_j = (a_{j1}, a_{j2}, \ldots, a_{jm}),
\end{math}
\\
which specifies the sensitivity of item \(j\) to each of the \(m\) latent traits. In other words, each component \(a_{jk}\) \\represents how strongly item \(j\) relates to latent dimension \(k\). For a single item, we set \(a_{jk}\) to be the same across all \\the thresholds of a polytomous (graded response or rating scale) model.
Each item also has threshold or difficulty \\parameters, denoted as 
\(\quad b_{jk},\)
which indicate the location along the latent dimension(s) where the item optimally\\ differentiates between respondents with different trait levels. Since we use polytomous models, multiple\\ thresholds \(b_{jk}\) are used to correspond to different response categories. 
}\\
\midrule
Model & \makecell[l]{
The probability that an individual \(i\) with latent trait vector \(\boldsymbol{\theta}_i\) responds correctly (or supports)\\ item \(j\) is modeled by a multidimensional logistic function:\\
\begin{math}
P(u_{ij} = 1 \mid \boldsymbol{\theta}_i, \boldsymbol{a}_j, d_j) = \frac{1}{1 + \exp\left[\left(\boldsymbol{a}_j' \boldsymbol{\theta}_i + d_j\right)\right]},
\end{math}\\
where \(\boldsymbol{a}_j\) is the discrimination vector for item \(j\), and \(d_j\) is the difficulty (threshold) parameter.\\
For polytomous (ordinal) responses, MIRT generalizes the unidimensional graded response \\model by estimating the probability of responding in each category as the difference between\\ adjacent category response functions. For item \(j\) with response categories \(k = 1, \ldots, K\),\\ the probability of responding in category \(k\) given latent traits \(\boldsymbol{\theta}\) is modeled as:\\
\begin{math}
P(Y_j = k \mid \boldsymbol{\theta}) = P(Y_j \geq k \mid \boldsymbol{\theta}) - P(Y_j \geq k+1 \mid \boldsymbol{\theta}),
\end{math}
\\
where each \(P(Y_j \geq k \mid \boldsymbol{\theta})\) is computed using a multidimensional logistic function involving the\\ discrimination vector \(\boldsymbol{a}_j\), latent trait vector \(\boldsymbol{\theta}\), and category threshold parameters \(b_{jk}\). This approach\\ captures the ordered nature of responses while simultaneously considering multiple latent dimensions.
} \\
\midrule
Estimation Algorithm & \makecell[l]{The learning of the item parameters in IRT is typically enabled through expectation-maximization algorithm. However, \\QMCEM is better than traditional EM for multidimensional IRT because it uses quasi-random (evenly sampled) sequences\\ to approximate high-dimensional integrals, rather than relying on random sampling or standard numerical methods. This \\approach provides more even coverage of the multidimensional latent trait space, which reduces variance in the integral\\ estimates needed for parameter updates. QMCEM typically converges faster and yields more accurate and stable parameter\\ estimates in multidimensional settings. Standard EM  used in single-factor IRT could be slower, less precise, and prone to\\ instability in high dimensions due to the inefficiency and unevenness of random samples used for the needed integrations.}\\
\bottomrule
\end{tabular}
}
\caption{Details on MIRT model that was used for the experiments.}
\label{tab:IRT}
\end{table*}

\section{Dataset Details}

\begin{figure}[!ht]
    \centering
\includegraphics[width=.9\linewidth]{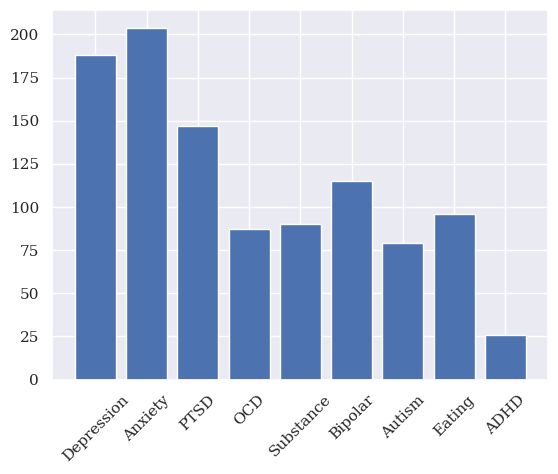}
    \caption{The number of participants in the dataset that reported diagnosis for each of the conditions. 
}
    \label{fig:dist_diagnoses}
\end{figure}

Figure~\ref{fig:dist_diagnoses} shows the distribution of diagnoses among all participants in the dataset, and Table~\ref{tab:questionnaire} lists the language-response questions. Because AUDIT and DUDIT assess overlapping symptoms, our language-based questions grouped alcohol and drug use under a broader substance abuse category, allowing participants to discuss their addictions more generally rather than focusing on alcohol or drugs in particular. Additionally, although ADHD is common in the general population, it is underrepresented in our dataset; many participants with an ADHD diagnosis dropped out before completion due to the survey’s overall length (over 100 questions).

\label{app:questionnaire}

% \usepackage{graphicx}
% \usepackage[normalem]{ulem}
\begin{table*}

\begin{tabular}{lp{0.75\linewidth}p{0.15\linewidth}}
\toprule
\textbf{ID} & \textbf{Question Text} & \textbf{Response Type} \\
\midrule
A1& \begin{tabular}[c]{@{}l@{}}Describe your worries and their strength, in the past few weeks.\\ Write 5 descriptive words. If this statement does not resonate with you,\\please type ‘not relevant’ in the first text box.\end{tabular} & words \\
A3 & \begin{tabular}[c]{@{}l@{}}Describe how your mood has influenced your behavior in the past few weeks.\\ Write at least 3 descriptive words.\end{tabular} & words \\
A4 & \begin{tabular}[c]{@{}l@{}}Describe places or activities you have avoided due to anxiety. Write at least\\ 3 descriptive words. If this statement does not resonate with you, \\please type ‘not relevant’ in the first text box.\end{tabular} & words \\
ADHD1 & \begin{tabular}[c]{@{}l@{}}Describe your attention during tasks or assignments. Think about your \\workplace or school. Write at least 3 descriptive words.\end{tabular} & words \\
ADHD2 & \begin{tabular}[c]{@{}l@{}}Describe activities of restlessness, impulsivity, and decisions you made \\without thinking it through. Write at least 3 descriptive words.\\ If this statement does not resonate with you, please\\ type ‘not relevant’ in the first text box.\end{tabular} & words \\
ASD2 & \begin{tabular}[c]{@{}l@{}}Describe your typical social interaction and the typical way of communication.\\ Write at least 2 descriptive words.\end{tabular} & words \\
ASD3 & \begin{tabular}[c]{@{}l@{}}Describe situations where you are intensely focused on specific topics or \\activities, to the exclusion of others. Write at least 3 descriptive words.\\ If this statement does not resonate with you, please type ‘not relevant’\\ in the first text box.\end{tabular} & words \\
ASD4 & \begin{tabular}[c]{@{}l@{}}Describe situations where your senses were particularly overwhelmed,\\ or distressed. Write at least 3 descriptive words. If this statement does not\\ resonate with you, or you do not commonly experience such situations, \\please type ‘not relevant’ in the first text box.\end{tabular} & words \\
ASD5 & \begin{tabular}[c]{@{}l@{}}Describe your daily routine in general terms, and \\feelings when this routine is changed. Write at least 3 descriptive words.\end{tabular} & words \\
ASD6 & \begin{tabular}[c]{@{}l@{}}Describe how you navigate, experience, and maintain social \\relationships. Write at least 3 descriptive words.\end{tabular} & words \\
BD2 & \begin{tabular}[c]{@{}l@{}}You experienced recurring cycle of mood swings, moving from highs to lows \\ and back again. If so, can you share a timeline of when you \\experienced episodes of elevated mood followed by depressive episodes? \\If this statement does not resonate with you, please type ‘not relevant’ \\in the text box.\end{tabular} & essay \\
BD3 & \begin{tabular}[c]{@{}l@{}}Describe impulsive or risky behaviors you have been engaged in lately.\\ Write at least 2 descriptive words. If this statement does not resonate with you, \\please type ‘not relevant’ in the first text box.\end{tabular} & words \\
ED1 & \begin{tabular}[c]{@{}l@{}}Describe your eating habits that differ from other people. Consider your last \\ week. Write at least 3 words. If this statement does not resonate with you, \\please type ‘not relevant’ in the first text box.\end{tabular} & words \\
ED2 & \begin{tabular}[c]{@{}l@{}}Describe your thoughts about food. Write at least 2 words.\end{tabular} & words \\
ED3 & \begin{tabular}[c]{@{}l@{}}Describe your thoughts about your weight, shape, or appearance.\\ Write at least 2 words.\end{tabular} & words \\
ED4 & \begin{tabular}[c]{@{}l@{}}Describe the control over your eating behavior and related feelings.\\ Write at least 1 word.\end{tabular} & words \\
ED5 & \begin{tabular}[c]{@{}l@{}}Describe behaviors and emotions you relate to food.\\ Write at least 1 word.\end{tabular} & words \\
ED6 & \begin{tabular}[c]{@{}l@{}}Describe the impact your eating behaviors have on your daily life and \\relationships. Write at least 1 word.\end{tabular} & words \\
\end{tabular}
\end{table*}
\begin{table*}
\begin{tabular}{lp{0.75\linewidth}p{0.15\linewidth}}
\toprule
\textbf{ID} & \textbf{Question Text} & \textbf{Response Type} \\
\midrule
G1 & Describe your mental health in a paragraph. Write at least 300 words. & essay \\
G10 & \begin{tabular}[c]{@{}l@{}}When did you first notice difficulties in relation to your mental health?\\ (open response)\end{tabular} & essay \\
G12 & \begin{tabular}[c]{@{}l@{}}Describe how your emotions and social relations have been influenced by \\your mental health. Write at least 3 descriptive words.\\ If this statement does not resonate with you, please type ‘not relevant’ \\in the first text box.\end{tabular} & words \\
G2 & \begin{tabular}[c]{@{}l@{}}Describe your mental health. Write 5 descriptive words.\end{tabular} & words \\
G3 & \begin{tabular}[c]{@{}l@{}}Describe how your mental health has influenced your behavior in \\the past few weeks. Write at least 2 descriptive words.\end{tabular} & words \\
G4 & \begin{tabular}[c]{@{}l@{}}Describe how your mental health has influenced your work performance\\ in the past few weeks. Write at least 2 descriptive words.\end{tabular} & words \\
G5 & \begin{tabular}[c]{@{}l@{}}Describe how your body felt in the past few weeks. Think about physical \\symptoms that have relevance for you. Write at least 3 descriptive words.\end{tabular} & words \\
G6 & \begin{tabular}[c]{@{}l@{}}Describe things you have been unable to do, concentrate on, make decisions on, \\or carry out due to your mental health. Write at least 3 descriptive words.\\ If this statement does not resonate with you, please type ‘not relevant’.\end{tabular} & words \\
G7 & \begin{tabular}[c]{@{}l@{}}Describe how your mood has influenced your daily life, in the past few weeks.\\ Write 3 descriptive words.\end{tabular} & words \\
G8 & \begin{tabular}[c]{@{}l@{}}Consider your main mental health symptoms, how long have you been \\experiencing them? (open response)\end{tabular} & essay \\
G9 & \begin{tabular}[c]{@{}l@{}}Describe how your attention and activity level have influenced your social\\ relationships. Write at least 3 descriptive words.\end{tabular} & words \\
G91 & Describe how your attention and activity level have influenced your work. Write at least 3 descriptive words. & words \\
OCD1 & \begin{tabular}[c]{@{}l@{}}Describe recurring thoughts you experienced, and their content, in the past\\ few weeks. Write at least 3 descriptive words.\\ If this statement does not resonate with you, please type ‘not relevant’ \\in the first text box.\end{tabular} & words \\
OCD2 & \begin{tabular}[c]{@{}l@{}}Describe actions or rituals that you felt compelled to perform repeatedly,\\ in the past few weeks. Write at least 3 descriptive words.\\  If this statement does not resonate with you,\\ please type ‘not relevant’ in the first text box.\end{tabular} & words \\
OCD3 & \begin{tabular}[c]{@{}l@{}}Describe obsessive thoughts or compulsions that you attempted to resist. \\ Write at least 3 descriptive words.\\ If this statement does not resonate with you, \\please type ‘not relevant’ in the first text box.\end{tabular} & words \\
OMD1 & \begin{tabular}[c]{@{}l@{}}Describe your changes, if any, in your mood or emotions in the past few weeks.\\ Write at least 2 descriptive words.\end{tabular} & words \\
OMD2 & \begin{tabular}[c]{@{}l@{}}Describe a persistent mood or emotions you experienced in the past few weeks.\\ Write 5 descriptive words.\end{tabular} & words \\
\end{tabular}
\end{table*}
\begin{table*}
\begin{tabular}{lp{0.75\linewidth}p{0.15\linewidth}}
\toprule
\textbf{ID} & \textbf{Question Text} & \textbf{Response Type} \\
\midrule
OMD3 & \begin{tabular}[c]{@{}l@{}}Describe your ability to enjoy things in the past few weeks.\\ Write at least 2 descriptive words.\end{tabular} & words \\
OMD4 & \begin{tabular}[c]{@{}l@{}}Describe how your appetite has been lately.\\ Write at least 1 descriptive word.\end{tabular} & words \\
OMD5 & \begin{tabular}[c]{@{}l@{}}Describe how your sleep has been lately.\\ Write at least 1 descriptive word.\end{tabular} & words \\
OMD6 & \begin{tabular}[c]{@{}l@{}}Describe how your motivation and/or energy level has been lately.\\ Write at least 2 descriptive words.\end{tabular} & words \\
PTSD1 & \begin{tabular}[c]{@{}l@{}}Describe impactful events you experienced and that are still \\influencing your life. Write a paragraph with at least 300 words.\end{tabular} & essay \\
PTSD2 & \begin{tabular}[c]{@{}l@{}}Describe impactful events you experienced and that are still\\ influencing your life. Write 5 descriptive words.\end{tabular} & words \\
PTSD3 & \begin{tabular}[c]{@{}l@{}}Describe thoughts, memories, or dreams related to impactful events that\\ are influencing your life. Write 5 descriptive words.\\ If this statement does not resonate with you, \\please type ‘not relevant’ in the first text box.\end{tabular} & words \\
PTSD4 & What was the traumatic event? (open response) & essay\\
SUB1 & \begin{tabular}[c]{@{}l@{}}List drugs or substances that you have used.\\ Include alcohol in this list, if relevant. (open response)\end{tabular} & essay \\
SUB2 & \begin{tabular}[c]{@{}l@{}}Describe the circumstances under which you use substances.\\ Write at least 2 words.\end{tabular} & words \\
SUB3 & \begin{tabular}[c]{@{}l@{}}Describe your thoughts, behavior, and feelings when you are\\ not using substances that you typically use.\\ Write at least 1 word.\end{tabular} & words \\
SUB4 & \begin{tabular}[c]{@{}l@{}}Describe social, educational, or occupational consequences you experienced\\ due to your usage of substances. Write at least 1 word.\end{tabular} & words \\
SUB5 & \begin{tabular}[c]{@{}l@{}}Describe risky behavior that you engage in during your usage of substances. \\Write at least 3 descriptive words. If this statement does not resonate with you,\\ please type ‘not relevant’ in the text box.\end{tabular} & words \\
SUB6 & \begin{tabular}[c]{@{}l@{}}Describe your tolerance level towards substances. Write at least 1 word.\end{tabular} & words \\

\bottomrule
\end{tabular}%

\caption{The language response question that the participants responded to, along with the type of response and question code. The relevance of the question codes to specific factors is shown in Table~\ref{tab:question-level}.}
\label{tab:questionnaire}
\end{table*}
%The list of questions eliciting language responses are given in Table~\ref{tab:questionnaire}.

% \begin{figure*}[!ht]
%     \centering  \includegraphics[width=0.8\linewidth]{latent_figs/mulirt_covergence_plots.pdf}
%     \caption{\vv{Appendix}The performance of multidimensional adaptive testing that is modeled on a two-factor structure. On the x-axis is the number of questions asked, and on the y-axis is the Pearson correlation between the estimated factor score obtained from IRT against the relevant clinical scale. }
%     \label{fig:mulirt-results}
% \end{figure*}

% \begin{figure*}
%     \centering
%     \includegraphics[width=\linewidth]{latent_figs/mulirt_stabilization_points.pdf}
%     \caption{\vv{Appendix}Stabilization points determined by rolling std dev}
%     \label{fig:enter-label}
% \end{figure*}

\section{Question-level factors loading}
\begin{figure}[!ht]
    \centering
    \includegraphics[width=.9\linewidth]{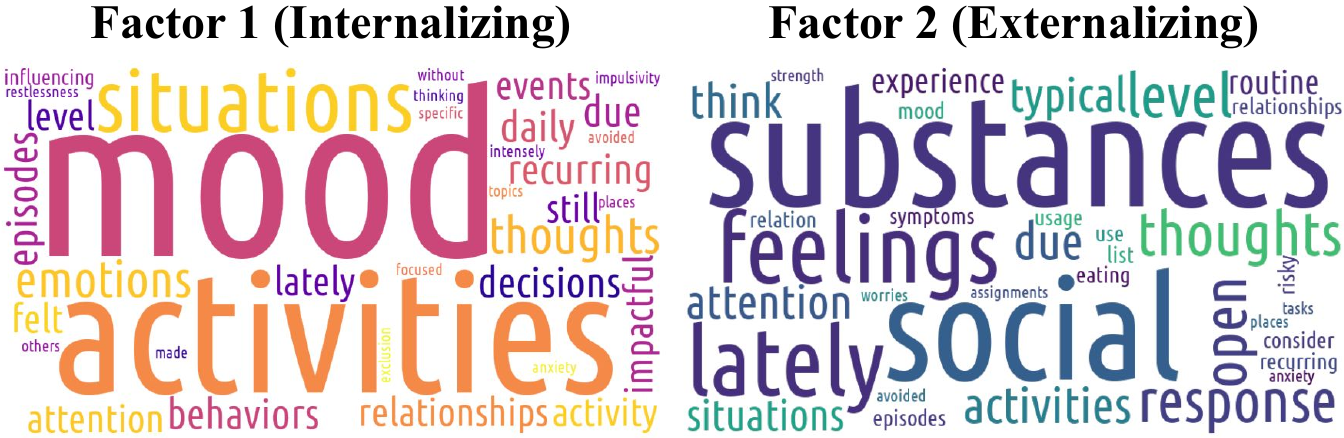}
    \caption{Question texts loading on to the two factors.}
        \label{fig:factor_words}
\end{figure}

After training the multi-outcome models, we found the multitask input aggregation model to perform the best. This model was then fed question-answer pair representations for all the training set as input, and the model inferred scores for each of the question-answer pair across all the users. These scores were then aggregated at a \textit{question-level} for applying the factor analysis model derived on user-aggregated scores, to understand how much each question contributes to understanding the two factors found to be significant, which is shown in Table~\ref{tab:question-level}, and visually shown as a wordcloud in Figure~\ref{fig:factor_words}. This was used to define the MIRT model for training and adaptive testing.

\section{ Stabilization Points}
\label{app:stabilization_points}
The stabilization points are given in a table along with the improvement over the random ordering and GPT-4 in Table~\ref{tab:condition_scores}. 
\begin{table}[!ht]
\centering
\resizebox{\columnwidth}{!}{
\begin{tabular}{lccccccl}
\toprule
\textbf{Condition} & \textbf{1} & \textbf{8} & \textbf{16} & \textbf{24} & \textbf{32} & \textbf{48} & \makecell{\textbf{Ques. to }\\\textbf{stabilize}}\\
\midrule
Depression \\
\hspace{2mm}GPT-4 & -.001 &.375&.518&.618&.637&.608&$>$ 48\\
\hspace{2mm}Random  & .327 & .642 & .686 & .759 & .753 %& .762
& .778 & 42  \\
\hspace{2mm}\textsc{MAQuA} & .324 & \textbf{.743 }& \textbf{.772} & \textbf{.782} & \textbf{.783} %& \textbf{.781}
& .781 &12 (71\%$\downarrow$)\\
\midrule
Anxiety \\
\hspace{2mm}GPT-4 & -.029 & .293 &.461 &.527 &.565 &.587&$>$ 48\\
\hspace{2mm}Random  & .332 & .612 & .647 & .690 & .694 %& .720
& .730 &34 \\
\hspace{2mm}\textsc{MAQuA} & .301 & \textbf{.650} & \textbf{.705} & \textbf{.716} & .\textbf{723} %& .726 
& .727 &13 (62\%$\downarrow$)\\
\midrule
ADHD\\ 
\hspace{2mm}GPT-4 & - &.248 & .419 & .447 &.513 &.486&$>$ 48\\
\hspace{2mm}Random  & .282 & .490 & .497 & .568 & .569 %& .563 
& .579 &41\\
\hspace{2mm}\textsc{MAQuA} & .235 & .482 & \textbf{.540} & .561 & .563 %& .570 
& .576 &17 (56\%$\downarrow$)\\
\midrule
Autism\\
\hspace{2mm}GPT-4 & -  &.180 & .324 &.441 &.496 &.502&$>$ 48 \\
\hspace{2mm}Random  & .184 & .320 & .298 & .361 & .354 %& .354 
& .360 &43\\
\hspace{2mm}\textsc{MAQuA} & .194 & .295 & .329 & .347 & .360 %& .363 
& .366 &6 (86\%$\downarrow$)\\
\midrule
Bipolar \\
\hspace{2mm}GPT-4 & -&.195& .323&.398 &.474 & .448&$>$ 48\\
\hspace{2mm}Random  & .202 & .420 & .390 & .452 & .454 %& .472 
& .473 & $>$ 48 \\
\hspace{2mm}\textsc{MAQuA} & .201 & \textbf{.440} & \textbf{.467} & \textbf{.470} &\textbf{ .472} 
%& .470 
& .473 &12 (75\%$\downarrow$)\\
\midrule
\multicolumn{2}{l}{Eating Disorder}\\
\hspace{2mm}GPT-4 & -  &.213 &.358 &.451 &.479 &.468&$>$ 48 \\
\hspace{2mm}Random  & .096 & .197 & .199 & .200 & .223 %& .268 
& .263 &$>$ 48\\
\hspace{2mm}\textsc{MAQuA} & .116 & \textbf{.218} &\textbf{ .221} & \textbf{.229} & .249 %& .254 
& .261 &7 (85\%$\downarrow$)\\
\midrule
OCD \\
\hspace{2mm}GPT-4 & -&.163 &.246&316 &.307&.336&$>$ 48\\
\hspace{2mm}Random  & .143 & .253 & .261 & .278 & .303 %& .310 
& .318 &39\\
\hspace{2mm}\textsc{MAQuA} & .120 & \textbf{.253 }& \textbf{.300} & \textbf{.305} & \textbf{.312} %& .312 
& .318 &19 (51\%$\downarrow$)\\
\midrule
\multicolumn{2}{l}{Alcohol Use}\\
\hspace{2mm}GPT-4 & - &.343 &.425 & .525 &.535 &.537&$>$ 48\\
\hspace{2mm}Random & .115 & .191 & .249 & .269 & .285 %& .303 
& .309 &40\\
\hspace{2mm}\textsc{MAQuA} & .105 &\textbf{ .320} & \textbf{.327} & \textbf{.311 }& \textbf{.301 } %& .303 
& .307 &5 (87\%$\downarrow$)\\
\midrule
PTSD\\
\hspace{2mm}GPT-4 & - & .220 & .330 & .407 &.438 &.424&$>$ 48\\
\hspace{2mm}Random  & .236 & .474 & .468 & .502 & .499 %& .526 
& .526 &35\\
\hspace{2mm}\textsc{MAQuA} & .176 & .464 & \textbf{.511} & \textbf{.513} & .517 %& .523 
& .527 &13 (63\%$\downarrow$)\\
\midrule
Drug Use\\
\hspace{2mm}GPT-4 & - & .240 & .489 &.530 &.516 &.521&$>$ 48\\
\hspace{2mm}Random  & .078 & .287 & .309 & .343 & .430 %& .433 
& .451 &$>$ 48\\
\hspace{2mm}\textsc{MAQuA} & \textbf{.114} & \textbf{.458} &\textbf{ .464} &\textbf{ .447} %& \textbf{.447} 
& \textbf{.445} & .447 &6 (87\%$\downarrow$)\\
\bottomrule
\end{tabular}}
\caption{Random and \textsc{MAQuA} Scores by condition across the 48 questions, each averaged across 20 runs. The last column shows the number of questions it takes for stabilization. }
\label{tab:condition_scores}
\end{table}

% \begin{figure}[!ht]
%     \centering
%     \includegraphics[width=\linewidth]{latent_figs/mulirt_average_corr_stabilizing.pdf}
%     \caption{The stabilization points for the correlations averaged across all the 10 mental health conditions. }
%     \label{fig:avg_stabilizing}
% \end{figure}

% \usepackage{graphicx}

% \usepackage{graphicx}
\begin{table*}[]
\resizebox{\textwidth}{!}{%
\begin{tabular}{ll}
\toprule
\makecell[l]{Carefully read the entire transcript, considering both explicit answers and patterns, emotional tone, and recurring themes throughout the conversation.\\         Rely exclusively on the transcript; do not infer information not present or speculate beyond what is written.\\         For each mental health dimension below, provide only the mental health scores requested, nothing more:\\ \\         * A single numerical score, adhering to the questionnaire's scale (see below for ranges).\\         * If insufficient information is present to make an informed rating for a dimension, give a score of 0.\\ \\         Mental health dimensions and corresponding standardized rating scales:\\ \\         * Depression score: PHQ-9 (range: 0–27)\\         * Anxiety score: GAD-7 (range: 0–21)\\         * Mood\_Disorder score: MDQ (range: 0–15)\\         * Obsessive\_Compulsive Disorder score: Y-BOCS (range: 0–40)\\         * Autism score: RAADS (range: 0–42)\\         * Eating\_Disorder score: EDE-QS (range: 0–72)\\         * PTSD score: NSESSS (range: 0–36)\\         * Alcohol\_Use score: AUDIT (range: 0–40)\\         * Substance\_Use score: DUDIT (range: 0–44)\\         * ADHD score: ASRS Part A (range: 0–24)}\\
\bottomrule
\end{tabular} 
}
\caption{Prompt for score estimation with Llama3.2 1B model.}
\label{tab:llama32_prompt}
\end{table*}

\begin{table*}[]
\resizebox{\textwidth}{!}{%
\begin{tabular}{l}
\toprule
\makecell[l]{You are an expert conversational agent conducting a mental health assessment. \\Your goal is to select, one at a time, the most informative question from a provided list to accurately score ten standardized mental health dimensions. \\Use prior conversation context to guide your choices. After each patient response, reassess and select the next optimal question.\\ If no further questions will improve scoring or none remain, output final scores for each dimension. \\Never answer the questions yourself—only select a question to ask or output the final scores.\\  \textless{}interview\_history\textgreater\\ \\ This section contains the interview history or conversational context so far. \\Use this information to inform your next question selection. \\If this is empty, you are at the start of the interview.\\ ...\\ \textless{}/interview\_history\textgreater\\ \\ \textless{}candidate\_questions\textgreater\\ This section contains the list of questions left to be asked at the current turn. \\Each question is labeled with a unique code and text. \\Select the most informative question from this list to ask next, unless you determine that the assessment should stop.\\ ...\\ \textless{}/candidate\_questions\textgreater\\ \\ Output format is as below. Only output in this JSON format with the mentioned keys, don't output anything else:\\  \{ \\question\_code: a single question code from candidate\_questions to ask next, or -1 if stopping, \\depression\_score: X/100, \\anxiety\_score: X/100, \\mood\_disorder\_score: X/100, \\ocd\_score: X/100, \\autism\_score: X/100, \\eating\_disorder\_score: X/100, \\ptsd\_score: X/100, \\alcohol\_use\_score: X/100, \\substance\_use\_score: X/100, \\adhd\_score:X/100\\\}}\\
\bottomrule
\end{tabular}
}
\caption{Adaptive Question-asking prompt for GPT-4}
\label{tab:gpt4_prompt}
\end{table*}

\end{document}